# Self-Organization of the Neuron Collective of Optimal Complexity

V.G. Schetinin and A.W. Kostunin

Penza State Technical University  Penza State Pedagogic University
40, Krasnaya, Penza, 440017, Russia  37, Lermontova, Penza, 440026, Russia
E-mail: pip@diamond.stup.ac.ru

**ABSTRACT**

The optimal complexity of neural networks is achieved when the self-organization principles is used to eliminate the contradictions existing in accordance with the K. Godel's theorem about incompleteness of the systems based on axiomatics. The principle of S. Beer's exterior addition the Heuristic Group Method of Data Handling by A. Ivakhnenko realized is used.

## 1. INTRODUCTION

In many cases, the neural networks must be synthesized on the unrepresentative learning set composed of the small number of the classified instances. The instructions of the "teacher" used to learn a neural network do not usually exhaust a multitude of all possible states. Therefore, a few causal relations can represent the learning set classified. That is, the relations between the neural network inputs and its output can be represented a few functions that belong to a selected class of the transformations. The growing of the neural network complexity estimated by the number of its possible states increases the number of these functions. However, the variety of the neural network uselessly to increase more than the some value. Similarly, the number of functions used to describe the input-output relationships must be limited. This number should accord to the W. Ashby's principle of the adequate variety.

On the other hand, the learned neural networks can be regarded as the causal systems based upon axiomatics. The learning set can belong to this kind of the initial statements composed of the "teacher" instructions. We can require that the learning set, as a set of axioms, would not included the contradictory instances. However, the functions whose necessity is not possible to prove or refuse within the accepted axiomatics can be included into the desired collective of functions describing the relations between neural network inputs and outputs. It follows from the fundamental theorem of K. Godel concerning incompleteness [1].

The functions, whose necessity is not possible to prove or refuse, can generate the contradictory decisions on the input values that are close to each other and not represented by the learning set. These contradictions can be discovered on a testing set consisting of the classified instances, which were not included into learning set. Under these conditions, the number of the errors the trained neural network generated on the learning set must be minimal.

Thus, a neural network can be represented as a collective of the neurons that describe the discovered relations between its inputs and output. The number of neurons in the collective equaled to the number of these relations must be minimal. Each neuron has its local field of competence where the number of its errors is minimal. The borders of these fields are determined on learning set. For input values that are close to the borders of fields, the results of the whole collective of neurons should be taken into account. The decisions of neurons should be weighted in accordingly with their efficiency, which has been estimated on learning set. When the efficiency of neurons is equal, the rule of majority votes is used in order to evaluate the plausibility of the taken decision. For example, the value of plausibility may be equal to the ratio of the number of the neurons voted for the taken decision to their total number.

The complexity of the learned neural network will be optimal if the number of its errors occurred on leaning set as well as the number of the neurons in the collective and the number of their inputs are minimal. Therefore, the number of the functions that are able to generate the contradictory decisions will also minimal.

The neural network of optimal complexity can be synthesized on the unrepresentative learning set using the fundamental principles of the self-organization [2]. In order to exclude the contradictory relations the trained neural network can represent, the principle of exterior addition of S. Beer would be applied [1, 2]. This principle the Heuristic Group Method of Data Handling (GMDH) of A. Ivakhnenko has been used to synthesize the neural networks consisting of an optimal number of layers [3, 4, 5].

In the many cases, the efficiency of the neural networks can be increased if instead a quantitative input variables to use their products as the generalized variables. The use such variables a nonlinear transformation formed may be more preferable than expanding the structure of the input variables [6]. Below, the methods for self-organizing the neural networks whose optimal complexity is achieved with a nonlinear transformations are discussed.

## 2. THE BASIC STAGES OF SELF-ORGANIZATION

The neural network behavior can be described by logical (Boolean) functions how McCaloch and Pitts have suggested [2]. Let the number of input variables $x_1, ..., x_m$ is $m$, and there exist $L$ logical functions $f_i(x_1, ..., x_m)$, $i= 1, ..., L$, where $L= 2^q$, $q= 2^m$. Note that when $m \geq 5$, the number $L > 10^{10}$ and searching for all $L$ the variants of logical functions of $m$ variables requires the huge computational expenses.

In order to reduce these expenses, the several methods were suggested. One of them is the method of D. Willis in which the initial logical function of many variables decomposes on the several functions of the smaller number of the arguments [2]. However, this method is applicable to the narrow class of logical functions. Another approach that has suggested based on the GMDH that is applicable to a wider class of functions [3, 4]. The heuristics of the HMDH has been used to synthesize the efficient neural networks on the unrepresentative learning set. Using the HMDH, the learning set may include even $n= 6$ instances the «teacher» classified as $y_i^\circ$, $i= 1, ..., n$.

Within GMDH, the neural network is represented as a multilayered scheme of F. Rosenblatt perceptron. It is synthesized with the reference function $g(v_1, ..., v_p)$ of $p$ arguments $v_1, ..., v_p$, usually $p= 2$. The reference function $g()$ may belong to an arbitrary class of functions (e.g., the class of Kolmogorov-Gabor polynomials and the logical functions).

Using the reference function $g()$ in each layer $r$, all the variants of function-candidates $f_i$ whose values noted $z_i$ are generated

$$z_i = g(u_1, u_2). \qquad (1)$$

One of the possible GMDH algorithms is the case

$$u_1 = z_j(r-1), j= 1, ..., F;$$
$$u_2 = x_k, k= 1, ..., m, \qquad (2)$$

where $F$ is the number of the function-candidates which are best on a criterion $CR$ in the layer $r$.

The number $L$ of function-candidates equals to $C_m^2 = m(m-1)/2$ in the first layer and $Fm$ in the next ones. Note that the number $F \cong 0.4L$. It means that $F$ best function-candidates are selected to be in the next layers. Using this, the D. Gabor's principle of the inconclusive decisions is realized in the GMDH [3, 4].

The criterion $CR$ used to select the function-candidate supposes that the learning set should be divided into two or more non-conjunctive subsets $A$, $B$, ... of same length. These subsets are used to learn the neural networks on each of them to realize a heuristics. For this heuristics, the true function $f^*$ that accurately describes the neural network behavior would not depend on the choice of the learning set subsets (i.e., subset $A$ or $B$). The efficiency of the function-candidate $f(W/I)$ synthesized on the subset $I= A, B, ...$ is estimated upon the set $W= A+ B+ ...$ Formally, this heuristic is expressed with the criteria of unbias $b_u$ and regularity $\Delta$:

$$b_u = |f(W/A) - f(W/B)|,$$
$$\Delta = |f(W/A) - Y^\circ| + |f(W/B) - Y^\circ|, \qquad (3)$$

where $Y^\circ = (y_1^\circ, ..., y_n^\circ)$ is the vector of instructions.

Since the efficiency of the function-candidate $f(W/I)$, $I= A, B, ...$ is estimated upon the instances belong to the subset $J \neq I$, the criterion $CR$ (3) is called exterior. The similar structure of criterion $CR$ needs to realize the principle of exterior addition introduced to exclude the above-mentioned contradictions.

For all function-candidates $f_i$ synthesized with the reference function (1), the values of criteria $b_u$ and $\Delta$ are computed in layers $r= 1, 2, ...$ In some GMDH algorithms, the convolution of these criteria is done. In all cases, the function-candidate in layer $r$ that is most efficient has the smallest value of the criterion $CR_m^{(r)}$.

While the value $CR_m^{(r)}$ in layers $r= 1, 2, ...$ is decreased, the complexity of the function $f_i^{(r)}$ is increased. Since the exterior addition principle is used, the value of the criterion goes through the minimum that points to the desired function $f^*$. However, because of the indeterminate components presenting in the input variables, the founded minimum may be local. To avoid the possible reducing the neural network efficiency, a positive variable $\delta > 0$ is introduced into stopping rule

$$CR_m^{(r-1)} \leq CR_m^{(r)} + \delta. \qquad (4)$$

Nevertheless, this rule is probably fulfilled for the desired function $f^*$. Under the noise or the distortion in the input variables, typically the rule (4) points to a function whose efficiency is lower.

Generally, the advantages of the GMDH algorithms are mostly concerned with the structure of the learned neural network. For self-organizing the neural network, we must not assign nor the number of layers nor the number of neurons as well as an activation function. Also, the learned neural network comprises such input variables called the features that are useful to separate the patterns of classified instances.

Thus, the GMDH algorithms are able to optimize the complexity of multi-layered neural networks.

## 3. THE MULTILAYERED SELF-ORGANIZATION

We have explored the GMDH-type algorithms of self-organizing and concluded that the structure and the synaptic weights of the learned neural network really depend on the following conditions. The results depend, firstly, on the variants of dividing the learning set into the subsets, secondly, on the choice of the number $F$ best function-candidates as well as on the value $\delta$ and, thirdly, on the choice of the exterior criterion structure [5, 7]. Also, when learning set is dividing on the two and more subsets, the undesirable uncertainty of synaptic weights increases.

The class of logical functions is more attractive for self-organizing the neural networks. The trained neural network may be easily interpreted as symbolic 'IF, THEN" rules to be used in expert systems or realized with the universal logical elements, etc.

For direct self-organizing the logical neural network of optimal complexity, the exterior criteria which have no above drawbacks were suggested [5, 8, 9]. In these criteria, the number $\nu$ of the neural network errors occurred on the learning set is used.

<u>Statement 1</u>. Let us $\nu_i$, $\nu_j$ and $\nu_k$ be the numbers of errors the function-candidates $f_i^{(r)}$, $f_j^{(r-1)}$ and input variable $x_k$ correspondingly generated. Note that $f_j^{(0)} = x_j$, and $j \neq k$, $k = 1, .., m$. Then it is sufficiently to select the functions $f_i^{(r)}$ for which

$$\nu_i < \min(\nu_j, \nu_k). \qquad (5)$$

If this condition is fulfilled, then the addition to the function $f_j^{(r-1)}$ the variable $x_k$ introduced will be exterior. Only such the addition makes the function $f_i^{(r)}$ more efficient.

Apparently, that while the number $r$ of layers increases, the number $\nu$ of the errors decreases to its minimum or zero. The number of layers increases until the properties of exterior addition are useful. Similarly to the condition (4), the following rule for stopping the algorithm may be formulated.

<u>Statement 2</u>. Let $L_r$ be the number of all the functions into $r$ layer for which the condition (5) is fulfilled. Then self-organizing the neural network should be stopped into $r = r^*$ layer if one of two conditions is fulfilled

$$CR_m^{(r)} = 0, \qquad (6)$$

$$L_{r+1} = 0.$$

When the second rule meets, i.e., $L_{r+1} = 0$, then the value $CR_m^{(r)} > 0$ to be. In this case, the new input variables should be added or the doubtful instances the trained neural network has erroneously classified should be excluded from the learning set.

When one of the conditions (6) is fulfilled, the number neurons that generate the equal number of errors may be more than one, i.e., $L_r > 0$. In this case, the collective of neurons can be made up. Each of these neurons has the same number of input variables.

Typically, the decisions the neurons generated are not coordinated with each other. The degree of their coherence may be estimated with the coefficient $\chi = n_1/N$, where $n_1$ is the number of the neurons voted for the taken decision; the $N$ is the total number of the neurons into the collective. The than closer the value $\chi$ to 1, the than more the coherence of the decisions. Usually, the value $\chi$ is maximal on the learning instances. Its value can be decreased upon the testing instances, which have not been included into learning set.

If the values of the coefficient $\chi$ are less than a value $\chi_0$, the decision of neuron collective is refused. Typically, the value $\chi_0$ is set no less than 0.8. Analyzing the values $\chi$ and $\chi_0$, the quality of learning set can be controlled. Apparently, when the value $\chi < \chi_0$, it is necessary either to involve the new input variables or to modify the learning set.

The values of the coefficient $\chi$ can be computed for the set of Boolean input variables. This set contains of $2^m$ combinations of the input variables. However, for realizing the logical neural network, the input quantitative variables must be quantized and represented as Boolean ones. The easiest manner to do it is to introduce the threshold $u_i$, i.e., the value which must be selected such that the quantized variable $x_i$ could generate the minimum number of the errors on the learning set. Apparently, such rough approximation will worsen the separating ability of the features.

For example, we can generate on the plain of two variables $x_1$ and $x_2$ a few points of two classes, which can be linearly separated. Parallel with these coordinates, we can draw the lines $x_1' = u_1$ and $x_2' = u_2$ indicating the quantized features. Obviously, these features are not able to exactly classify the all points of the set. We would try to improve their separating ability with transforming the input variables before the their quantization.

## 4. THE NONLINEAR TRANSFORMATION

In order to solve this problem, we have suggested using a nonlinear transformation [5]. Among of a nonlinear function, we have chosen the non-parametric ones due to two reasons. First, the non-parametric functions are more robust to uncertainty the unrepresentative learning set comprised, and second the formed features are easily interpreted. The suggested transformation is analogously to the geometric average function $h_p(v_1, .., v_p)$ of the $p \geq 2$ arguments

$$y = h_p(x_i, .., x_j), i \neq j = 1, .., m, \qquad (7)$$

where $y$ is the generalized variable.

As the variable $y$ is later quantized, the function $h_p()$ may be simplified to be represented as the product of the input variables

$$y = x_i \otimes ... x_j. \qquad (8)$$

Below, the following statement is formulated.

<u>Statement.</u> Let the numbers $\nu_i, .., \nu_j$ of the errors the quantized variables $x_i, .., x_j$ generated be known. Then, the variable $y$ may be as generalized one, if the following unequality is fulfilled

$$\nu < \min(\nu_i, .., \nu_j), \qquad (9)$$

where $\nu$ is the number of the errors the variable $y$ gener-ated on the learning set.

The number $p$ of the co-factors in equality (8) should increase until the condition (9) is fulfilled. Note that the number of all their combinations equals to $2^m - 1$.

When the condition (9) is met, the input variables $x_i$, ..., $x_j$, can be substituted for the corresponding generalized variable $y$. It leads to the following results.

1. The number of the neurons in the collective can be decreased.

2. The number $\nu$ of the errors the neural network generated on learning set can be decreased.

Corollary 1. The coherence of the decisions the trained neural network produced may be increased with increasing the number $p$ co-factors the generalized variable $y$ contained.

Corollary 2. Since the condition (9) is met under the minimal value $p$, the nonlinear power of the generalized variable $y$ to be least.

## 5. THE RESULTS OF MODELING

The suggested method and criteria of self-organization were used in the clinical laboratory diagnostics of pathologies [5, 8, 9]. The neural networks of logical type were used because of they can essentially facilitate the clinical interpretation of the discovered rules.

The efficiency of the suggested method may be demonstrated on the simple example for recognizing the human sex when we know size $x_1$ and weight $x_2$ of 7 females and 7 males. These two classes of the instances were linearly separable therefor no errors of the recognition to be.

To apply the neural network of the logical type, it is required to quantize the variables $x_1$, $x_2$ and find the corresponding thresholds $u_1$, $u_2$. Note that, in particular, these variables do not ensure the unerring recognition. What will happen if we transform the quantitative variables $x_1$ and $x_2$ to the generalized variable $y$. We can see, that the condition (9) is met for the generalized variable $y = x_1 x_2$. The quantized variable $y$ is able to reduce the number of neurons from 2 to 1 and decrease the number of errors on learning set to 0. Thus, the transformation used in this example is able to raise the efficiency of the neural network.

## 6. CONCLUSION

The suggested method and criteria of self-organization are able to synthesize the logical neural network of the optimal complexity. The neural networks of this type have advantages since their decisions are more robust and easily interpreted. However, the quantization of the quantitative input variables can reduce the efficiency of the trained neural network. The suggested nonlinear transformation of these variables can compensate the efficiency decrease of the logical neural network.

The suggested method of self-organization may use in the systems of the medical diagnostics, the knowledge extraction systems, etc.